\documentclass[runningheads]{llncs}

\usepackage{eccv}

\usepackage{eccvabbrv}

\usepackage{graphicx}
\usepackage{tabularx}
\usepackage{booktabs}

\usepackage[accsupp]{axessibility}  

\usepackage{multirow}
\usepackage{threeparttable}
\usepackage[dvipsnames]{xcolor}

\definecolor{myred}{RGB}{120, 0, 0}
\definecolor{mygreen}{RGB}{0, 230, 0}

\newcommand\blfootnote[1]{%
  \begingroup
  \renewcommand\thefootnote{}\footnote{#1}%
  \addtocounter{footnote}{-1}%
  \endgroup
}

\usepackage{hyperref}

\usepackage{orcidlink}

\begin{document}

\title{SparseLIF: High-Performance Sparse LiDAR-Camera Fusion for 3D Object Detection} 

\titlerunning{SparseLIF}

\author{Hongcheng Zhang\inst{1}$^\ast$ \and
Liu Liang\inst{1}$^\ast$ \and
Pengxin Zeng\inst{1,2}$^\ast$ \and
\\Xiao Song\inst{1}$^\dagger$\and
Zhe Wang\inst{1}}

\authorrunning{Zhang et al.}

\institute{SenseTime Research \and
College of Computer Science, Sichuan University \\
\email{\{zhanghongcheng,liangliu1\}@senseauto.com}, \email{zengpengxin.gm@gmail.com}, \email{\{songxiao,wangzhe\}@senseauto.com}}

\maketitle
\blfootnote{$^\ast$ Equal Contribution}
 
\begin{abstract}
Sparse 3D detectors have received significant attention since the query-based paradigm embraces low latency without explicit dense BEV feature construction. However, these detectors achieve worse performance than their dense counterparts. In this paper, we find the key to bridging the performance gap is to enhance the awareness of rich representations in two modalities. Here, we present a high-performance fully sparse detector for end-to-end multi-modality 3D object detection. The detector, termed SparseLIF, contains three key designs, which are (1) Perspective-Aware Query Generation (PAQG) to generate high-quality 3D queries with perspective priors, (2) RoI-Aware Sampling (RIAS) to further refine prior queries by sampling RoI features from each modality, (3) Uncertainty-Aware Fusion (UAF) to precisely quantify the uncertainty of each sensor modality and adaptively conduct final multi-modality fusion, thus achieving great robustness against sensor noises. By the time of paper submission, SparseLIF achieves state-of-the-art performance on the nuScenes dataset, ranking \textbf{\emph{1st}} on both validation set and test benchmark, outperforming all state-of-the-art 3D object detectors by a notable margin.

\end{abstract}

\section{Introduction} \label{sec.Introduction}
LiDAR-camera-based 3D detection is essential for accurate and robust autonomous driving systems. The two modalities naturally provide complementary information, \ie, the camera offers high-resolution semantic information while LiDAR provides accurate geometric information. Therefore, camera and LiDAR sensors have been simultaneously deployed for reliable 3D object detection.

Various approaches have been proposed to thoroughly explore the compensating information in LiDAR and camera modalities. Conventional multi-modality 3D object detection approaches typically transform two modalities into a unified space for feature fusion. For example, PointPainting~\cite{vora2020pointpainting} and its variants~\cite{wang2021pointaugmenting, yin2021multimodal, xu2021fusionpainting} decorate raw point clouds with image pixel features. BEVFusion~\cite{liu2023bevfusion,liang2022bevfusion} transforms image view features into dense BEV space to fuse with LiDAR features. The dense paradigm has achieved remarkable success in recent years but suffers from cumbersome view transformation, resulting in high latency, limited detection distance, and limited upper-bound performance. Recent works introduce a sparse query-based paradigm without explicit view transformation. Some pioneering sparse detectors aggregate multi-modality features in one~\cite{yan2023cross,wang2023unitr} or two~\cite{bai2022transfusion} stages using global attention. However, the exhaustive global attention buries the advantages of the sparse paradigm and makes it difficult to benefit from long-term temporal information. Lately, a stream of works explores the fully sparse paradigm, which is free from the usage of global attention and dense BEV queries. For example, works like FUTR3D~\cite{chen2023futr3d} and DeepInteraction~\cite{yang2022deepinteraction} sample features from two modalities using reference points. Despite the huge advances, these methods still lag behind their dense counterparts. Thus, whether fully sparse multi-modality detectors can achieve superior performance compared to dense detectors remains an open question.

This paper presents SparseLIF, a high-performance fully sparse multi-modality 3D object detector that outperforms all other dense counterparts and sparse detectors. SparseLIF bridges the performance gap by enhancing the awareness of rich LiDAR and camera representations in three aspects, \ie, query generation, feature sampling and multi-modality fusion. \textbf{First}, we argue that the convention~\cite{wang2022detr3d}, which randomly generates queries, will suffer from extra efforts in learning to move the query proposals towards ground-truth targets. Here, we propose the Perspective-Aware Query Generation (PAQG) module to ease learning. In particular, PAQG injects a lightweight perspective detector composed of the coupled 2D and monocular-3D sub-networks on image features to predict and transform top-scored 3D proposals into query proposals. These input-dependent proposals will narrow the learning path toward ground-truth targets, thus enhancing the awareness of rich contexts in high-resolution images. \textbf{Second}, these queries with perspective priors will interact with features from two modalities via the RoI-Aware Sampling (RIAS) module. Instead of resorting to cumbersome global attention, 
the module locates the region of interest and then sample complementary features at merely several reference points under the guidance of prior queries, thus conforming to the fully sparse paradigm and enjoying low latency. \textbf{Third}, we observe that in realistic scenarios, LiDAR and camera usually suffer from various sensor problems as shown in \cref{fig.UAF}, which will make sensor inputs unreliable and uncertain, thus degrading the performance of multi-modality detectors. Hence, we propose the Uncertainty-Aware Fusion (UAF) module to precisely quantify the uncertainty of each modality and guide our model to focus on the trustworthy modality in multi-modality fusion, thus achieving great robustness against sensor noises. Our contributions are summarized as follows.
\begin{itemize}
    \item We point out that the key to bridging the performance gap between sparse detectors and their dense counterparts is to enhance the awareness of rich representations from LiDAR and camera feature spaces in three aspects, \ie, query generation, feature sampling, and multi-modality fusion.
    \item We present a high-performance fully sparse detector for LiDAR-camera-based 3D object detection. The proposed framework contains three key designs: (1) Perspective-Aware Query Generation (PAQG), which enhances the perspective awareness of query proposals on rich contexts in high-resolution images; (2) RoI-Aware Sampling (RIAS), which effectively refine prior queries by sampling complementary RoI features across two modalities; (3) Uncertainty-Aware Fusion (UAF), which conducts final multi-modality fusion under the guidance of quantified modality uncertainty.
    \item We conduct comprehensive experiments to demonstrate the effectiveness of our proposed method. As can be seen, SparseLIF outperforms all state-of-the-art 3D object detectors on the nuScenes dataset, ranking \textbf{\emph{1st}} on both the validation set and test benchmark.
\end{itemize}


\section{Related Work} \label{sec.Related Work}
This section briefly reviews the most related works on three topics: LiDAR-, Camera- and LiDAR-Camera-based 3D object detection.
\subsection{LiDAR-based 3D Object Detection}
LiDAR provides accurate geometric information, attracting much attention for single-modality 3D detection. Earlier methods~\cite{li2021lidar,qi2018frustum,qi2017pointnet,qi2017pointnet++, shi2019pointrcnn,yang20203dssd,chen2019fast} directly extract features from raw point clouds to predict 3D bounding boxes, but suffers from the complexity when processing large-scale point clouds. Modern approaches transform unordered points into structured formats such as range-view maps~\cite{li2016vehicle,meyer2019lasernet,fan2021rangedet, liang2020rangercnn,sun2021rsn,bewley2020range}, pillars~\cite{lang2019pointpillars}, voxels~\cite{zhou2018voxelnet,yan2018second,deng2021voxel,shi2020pv}. Then, main-stream approaches apply 2D/3D convolution-based head~\cite{lang2019pointpillars,zhou2019objects,yin2021center} to predict 3D bounding boxes. Inspired by the huge success made by transformers, some recent works adopt transformer blocks in feature encoder~\cite{mao2021voxel,sheng2021improving,yuan2021temporal} and 3D detection head~\cite{bai2022transfusion}.

\subsection{Camera-based 3D Object Detection}
Camera-based 3D object detection~\cite{huang2021bevdet,wang2022detr3d,liu2022petr,huang2022bevdet4d,li2023bevdepth,li2022bevformer,park2022time} has witnessed remarkable progress over the past few years since camera-based approaches have lower deployment cost compared with the LiDAR-based counterparts. 

Inspired by the huge success made by LiDAR-based 3D detection methods, Pseudo-LiDAR~\cite{wang2019pseudo} transforms images into pseudo-LiDAR point clouds via depth estimation, then conducts 3D object detection on those pseudo points with LiDAR-based approaches. 
A line of works (\eg DD3D~\cite{park2021pseudo}, FCOS3D~\cite{wang2021fcos3d} and CenterNet~\cite{zhou2019objects}) further propose end-to-end, single stage 3D object detectors by attaching extra 3D bounding box regression head to 2D detector. Those methods attempt to explicitly estimate depth to assist in 3D detection but show limited performance due to inaccurate depth estimation.

To implicitly incorporate depth information, another line of works~\cite{roddick2018orthographic,reading2021categorical,park2022time} perform 3D detection in BEV space. LSS~\cite{philion2020lift} predicts the categorical depth distribution for each pixel to lift pixel features into a frustum, then splats all frustums into BEV grids. Based on LSS~\cite{philion2020lift}, BEVDet~\cite{huang2021bevdet} and BEVDepth~\cite{li2023bevdepth} substantially boost performance. 
Inspired by transformer, BEVformer~\cite{li2022bevformer, yang2023bevformer} and VideoBEV~\cite{han2023exploring} directly extract spatial features from camera views using cross-attention. Without reliance on depth information, the explicit construction of dense BEV features still limits inference speed and effective detection distance. 

Another stream of works~\cite{jiang2023far3d} employ a top-down manner that does not suffer from the explicit construction of dense BEV features. Inspired by DETR~\cite{carion2020end}, DETR3D~\cite{wang2022detr3d} manipulates predictions directly in 3D space by indexing 2D features with a sparse set of 3D object queries. PETR~\cite{liu2022petr} further eases the overhead of the indexing operation. PETRv2~\cite{liu2023petrv2} and Stream PETR~\cite{wang2023exploring} utilize the temporal information of previous frames to boost 3D object detection but adopt the global cross attention, which is computationally expensive. 
Sparse4D~\cite{lin2022sparse4d,lin2023sparse4d,lin2023sparse4dv3} and SparseBEV~\cite{liu2023sparsebev} sparsely sample multi-frame/view/scale features for 4D reference points then fuse hierarchically, thus achieving 3D detection without relying on dense view transformation and global attention.

\vspace{-0.25cm}
\subsection{LiDAR-Camera-based 3D Object Detection}

Recently, LiDAR-Camera-based 3D detection~\cite{chen2017multi} has achieved great success in leveraging semantic and geometric information to reach impressive performance. Early approaches~\cite{sindagi2019mvx, vora2020pointpainting, wang2021pointaugmenting, huang2020epnet, yin2021multimodal} decorate raw point clouds with image features but compromise rich context information. FrustumPointNet~\cite{qi2018frustum}, FrustumConvNet~\cite{wang2019frustum}, and CenterFusion~\cite{nabati2021centerfusion} lift image proposals into 3D frustums with explicit depth estimation but show limited performance due to depth inaccuracy. 

Lately, motivated by LSS~\cite{philion2020lift}, BEVFusion~\cite{liu2023bevfusion,liang2022bevfusion} ease the reliance on depth estimation by projecting fine-grained image features into BEV space then conducting fusion with LiDAR features. AutoAlign~\cite{chen2022autoalign, chen2022autoalignv2} further preserves instance-wise semantic consistency by feature alignment across two modalities. However, the explicit and dense view transformation from image to BEV space is cumbersome (\ie, high latency and limited detection distance) and sensitive to sensor misalignment. BEVFusion4D~\cite{cai2023bevfusion4d} further improves performance by incorporating temporal information. EA-LSS~\cite{hu2023ea} enhances depth estimation at the edge of objects. 

Recent works utilize the sparse query-based paradigm without explicit view transformation. Transfusion~\cite{tian2019fcos} obtains object queries from LiDAR points and then fuses queries with rich image features using a transformer block. CMT~\cite{yan2023cross} further develops an end-to-end feature interaction framework for multi-modality fusion. UniTR~\cite{wang2023unitr} introduces a modality-agnostic transformer encoder to proceed with unified modeling and shared parameters. Despite great success, the expensive global attention buries the advantages of the sparse paradigm and makes it difficult to benefit from long-term temporal information.

Another stream of works explores the fully sparse paradigm. SparseFusion~\cite{zhou2023sparsefusion} poses detectors on each modality and fuses features of detected instances. However, the two-stage paradigm suffers from the limited performance of modality-specific detectors. FUTR3D~\cite{chen2023futr3d} generalizes the fully sparse paradigm by initializing 3D reference points and projecting them into all available modalities to sample features. Although methods have recently achieved good performance, there remains notable performance gap compared to dense counterparts. 


\begin{figure}[tb]
  \centering
  \includegraphics[width=\linewidth]{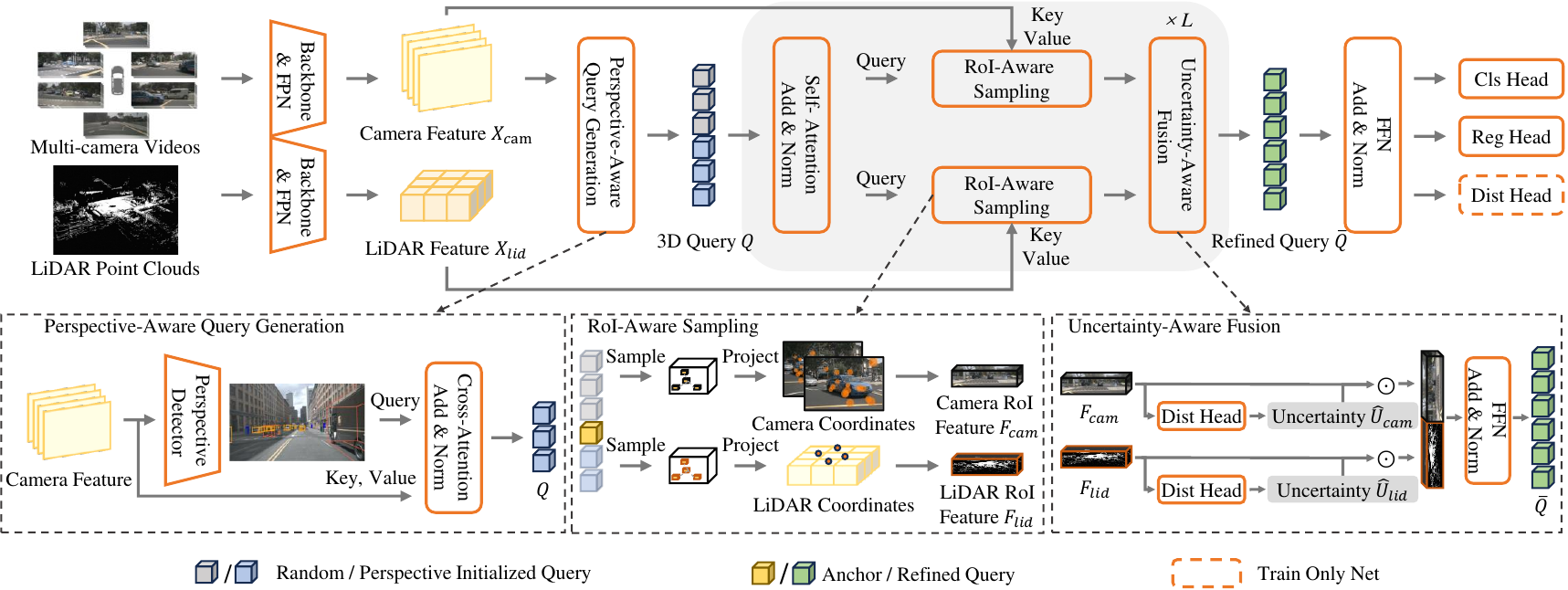}
  \caption{The overall architecture of SparseLIF, a fully sparse LiDAR-camera-based 3D object detector. The framework contains a camera backbone to process multi-view videos and a LiDAR backbone to encode raw point clouds. We then feed the image features into the Perspective-Aware Query Generation (PAQG) module to generate queries. The queries will interact with the camera and LiDAR features via the RoI-Aware Sampling (RIAS) module to extract complementary features for further refinement. Next, the Uncertainty-Aware Fusion (UAF) module quantifies the uncertainty of RoI features from two modalities and adaptively conducts final multi-modality fusion. The decoder repeats $L$ times. }
  \vspace{-0.1cm}\vspace{-0.1cm}
  \label{fig.framework}
\end{figure}
\section{SparseLIF} 

\label{sec.Method}
SparseLIF is a sparse query-based multi-modality detector. We use common image backbone (\eg ResNet~\cite{he2016deep}, V2-99~\cite{lee2020centermask}) and FPN~\cite{lin2017feature} to extract multi-view/scale/frame camera features, denoted as $X_{\text{cam}} = \{\mathcal{X}_{\text{cam}}^{vmt}\}_{v=1,m=1,t=1}^{V,M,T}$, where $V$, $M$, and $T$ denote the number of camera views, feature scales, and temporal frames respectively. Based on our proposed framework, rich temporal information can be easily and sufficiently incorporated. In parallel, we use common 3D LiDAR backbone (\eg VoxelNet~\cite{zhou2018voxelnet}) and FPN~\cite{lin2017feature}  to extract multi-scale LiDAR features, denoted as $X_{\text{lid}} = \{\mathcal{X}_{\text{lid}}^{r}\}_{r=1}^{R}$, where $R$ denotes the number of LiDAR feature scales. Taking camera features as input, the Perspective-Aware Query Generation (PAQG) module (\cref{sec.Perspective-Aware Query Generation}) adopts the coupled 2D and monocular-3D image detectors to predict and generate high-quality 3D queries with perspective priors. These queries will then interact with the camera and LiDAR features via the RoI-Aware Sampling (RIAS) module to extract RoI features for further refinement. Next, the Uncertainty-Aware Fusion (UAF) module (\cref{sec.Uncertainty-Aware Fusion}) quantifies the uncertainty of RoI features from two modalities and adaptively conducts multi-modality fusion for final 3D object predictions.

\subsection{Perspective-Aware Query Generation} \label{sec.Perspective-Aware Query Generation}

\begin{figure}[tb]
  \centering
  \includegraphics[width=\linewidth]{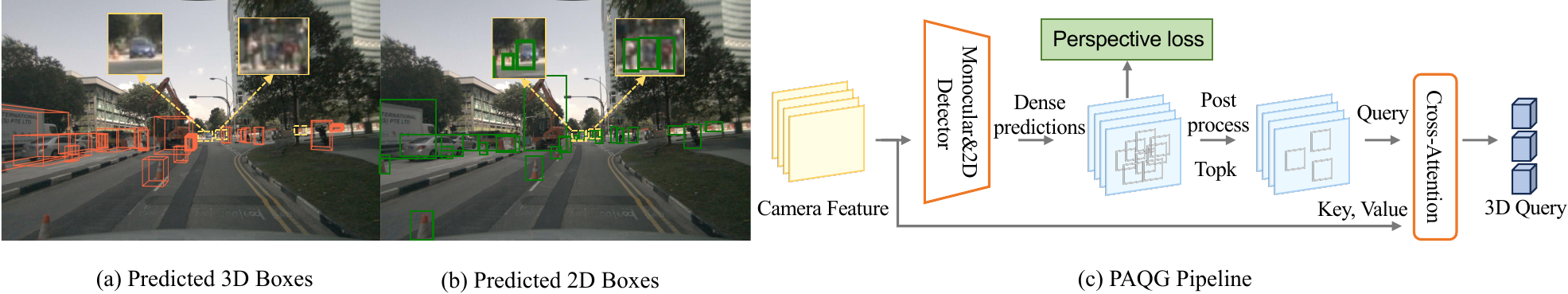}
  \caption{Motivations and details of our proposed PAQG module. \textbf{(a)} 3D detectors struggle with low sensitivity when detecting distant and small objects.  \textbf{(b)} 2D detectors demonstrate excellent pixel-wise perception capabilities on such objects.
  \textbf{(c)} the PAQG module adopts the coupled 2D and monocular-3D sub-networks to predict dense boxes under the supervision of a perspective loss. 
  We pick top-ranked boxes to propose high-quality queries, and then interact with camera features via a cross-attention module.
  }
  \label{fig.PAQG}
\end{figure}

Recent works typically generate queries based on randomly distributed reference points~\cite{chen2023futr3d, yan2023cross}, anchor boxes~\cite{lin2022sparse4d} or pillars~\cite{liu2023sparsebev} in 3D space and optimize as net parameters, regardless of input data. However, it has already been proved in 2D detection~\cite{yao2021efficient} that such input-independent queries will take extra effort in learning to move the query proposals towards ground-truth object targets. As shown in \cref{fig.PAQG}, we visualize the predictions of a query-based 3D detector and a 2D detector, where the 2D detector usually exhibits excellent perception capability on distant and small objects. Motivated by the strength of 2D detection, our PAQG module fully utilizes the perception capability to generate 3D queries, thereby assisting ultimate 3D detection.

The lightweight perspective detector in the PAQG module consists of the coupled 2D (\eg FCOS~\cite{tian2019fcos}) and monocular-3D (\eg FCOS3D~\cite{wang2021fcos3d}) sub-networks. Taken the multi-view/scale image features $X_{\text{cam}}$ as input, the monocular-3D sub-network predicts raw 3D attributes, \ie, depths $\mathbf{d}$, rotation angles, sizes, and velocities throughout different views. Simultaneously, the 2D sub-network predicts corresponding 2D attributes, \ie, center coordinates $[\mathbf{c_x},\mathbf{c_y}]$, confidence scores, and category labels. 
For each view $v$, we project the box centers into 3D space based on corresponding camera extrinsic ${E_v}$ and intrinsic ${I_v}$, \ie,
\begin{equation}
  \mathbf{c}^{3D} =  {E_v^{-1}}{I_v^{-1}}[\mathbf{c_x d},\mathbf{c_y d},\mathbf{d},\mathbf{1} ].
\end{equation}
The 3D center $\mathbf{c}^{3D}$ will combine with the predicted size, rotation angle, and velocity to form 3D boxes.
Then, we perform non-maximum suppression in 3D space to filter intersecting boxes and pick the top $N_k$ boxes ranked by confidence scores, to initialize queries with image features interacted via an efficient cross-attention module. Formally, 
\begin{equation}
  {q}_{i} = \frac{1}{|\mathcal{V}|} \sum_{v\in \mathcal{V}}\sum_{m=1}^{M}\mathcal{BS}(\mathcal{X}_{\text{cam}}^{vm},\mathcal{P}_{\text{cam}}^{v}(c^{3D}_i)),
\end{equation}
where $\mathcal{P}_{\text{cam}}^{v}(c^{3D}_i)$ projects the 3D center $c^{3D}_i$ to $v$-th image using corresponding camera parameters. Besides, $\mathcal{V}$ denotes the set of hit views. $\mathcal{BS(\cdot)}$ denotes the bilinear sampling function. Since some objects may be overlooked, we preserve $N_r$ randomly initialized query boxes. Finally, our PAQG module generates total $N_q = N_k + N_r$ query proposals. This way, our PAQG module provides input-dependent query proposals to elevate the understanding of comprehensive perspective priors (2D and 3D attributes) for 3D detectors, thereby aiding in detecting distant and small objects.

\subsection{RoI-Aware Sampling} \label{sec.RoI-Aware Sampling}
RoI-Aware Sampling (RIAS) module is responsible for sampling RoI features from each modality to refine the queries $Q=\{q_i\}_{i=1}^{N_q} \subset \mathbb{R}^{C}$  initialized with perspective priors via PAQG module. We aim at locating the region of interest (RoI) to sample features without resorting to cumbersome global attention, thus enjoying low complexity and benefiting from long-term temporal information. 

\subsubsection{LiDAR Branch} Inspaired by Deformable Attention~\cite{zhu2020deformable}, we merely sample $K=4$ reference points to retrieval RoI features $\{F_{\text{lid}}^{ik}\}_{k=1}^{K}$ from LiDAR feature map $\mathcal{X}_{\text{lid}}$ for each query $q_i$. Formally, 
\begin{equation}
  F_{\text{lid}}^{ik} = \sum_{r=1}^{R} \mathcal{BS} \left(\mathcal{X}_{\text{lid}}^{r},\mathcal{P}_{\text{lid}}\left(c_i + \Delta_{\text{lid}}^{irk}\right)\right)\cdot \sigma_{\text{lid}}^{irk},  
\end{equation}
where $c_i$ is the bounding box center of query $q_i$ in global 3D space and $\mathcal{P}_{\text{lid}}$ projects the center into LiDAR BEV space. $\mathcal{BS(\cdot)}$ denotes the bilinear sampling function. Besides, $\Delta_{\text{lid}}^{irk}$ and $\sigma_{\text{lid}}^{irk}$ are predicted sampling offsets and attention weights using query $q_i$ to cover the RoI on sensitive objects. Note that, different from global attention, we merely interact with several features mapped to reference points, thus embracing a fully sparse paradigm.

\subsubsection{Camera Branch}
As for the camera branch, we also sample $K=4$ reference points to retrieval RoI features from the hit views $\mathcal{V}$ of camera feature map $\mathcal{X}_{\text{cam}}$, \ie, 
\begin{equation}
  F_{\text{cam}}^{itk} = \frac{1}{|\mathcal{V}|} \sum_{v\in \mathcal{V}}\sum_{m=1}^{M} \mathcal{BS} \left(\mathcal{X}_{\text{cam}}^{vmt},\mathcal{P}_{\text{cam}}^{vt}\left(c_i + \Delta_{\text{cam}}^{ivmtk}\right)\right)\cdot \sigma_{\text{cam}}^{ivmtk},
\end{equation}
where $\mathcal{P}_{\text{cam}}^{vt}(\cdot)$ is the project function from global 3D space to feature coordinate using camera parameters and temporal alignment~\cite{liu2023sparsebev}.  Besides, $\Delta_{\text{cam}}^{ivmtk}$ and $\sigma_{\text{cam}}^{ivmtk}$ are also predicted sampling offsets and attention weights using query feature. 

\subsubsection{Channel-Spatial Correlation Aware Mixing}
To enrich the awareness of the correlation in spatial and channel dimensions of query $q_i$, we inject AdaMixer~\cite{gao2022adamixer} on the retrieved features.  For convenience, we organize those retrieved RoI features to $f \in \mathbb{R}^{S\times C}$, where $S = K$ or $S = T\times K$ for LiDAR or camera feature. 

First, we model the channel correlation based on query $q_i$ and transform features $f$ to enhance channel semantics:
\begin{align}
W_c & =\operatorname{Linear}(q_i) \in \mathbb{R}^{C \times C} \\
\operatorname{M}_c({f}) & =\operatorname{ReLU}\left(\text {LayerNorm}\left({f} W_c\right)\right),
\end{align}
where $W_c$ is the channel correlation shared across different timestamps and different sampling points. Next, we then transpose the feature and model the spatial correlation to the spatial dimension of it, \ie,
\begin{align}
W_s & =\operatorname{Linear}(q_i) \in \mathbb{R}^{S \times S} \\
\operatorname{M}_s({f}) & =\operatorname{ReLU}\left(\text {LayerNorm}\left({f}^T W_s\right)\right),
\end{align}
where $W_s$ is the spatial correlation shared across different channels. After channel-spatial correlation aware mixing, the features are flattened and aggregated by a linear layer. 

\begin{figure}[tb]
  \centering
  \includegraphics[width=\linewidth]{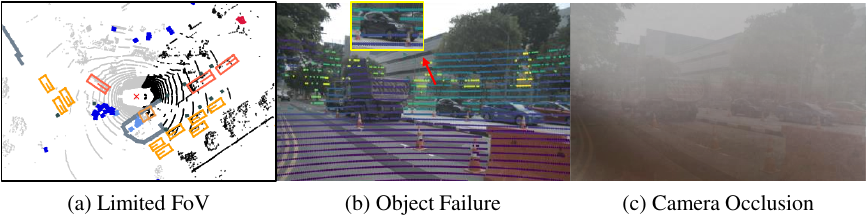}
  \caption{Visualizations of sensor noises in 3D object detection for autonomous driving. 
  \textbf{(a) Limited FOV}: LiDAR installed in a front-facing manner yields a limited FOV, \eg $120^{\circ}$. 
    \textbf{(b) Object Failure}: the reflection rate of some objects (\eg the black car) is below the threshold of LiDAR thus without LiDAR points reflected. 
    \textbf{(c) Camera Occlusion}: the camera module is usually vulnerable to occlusions (\eg by dust). 
    }
  \label{fig.UAF}
\end{figure}
\subsection{Uncertainty-Aware Fusion} \label{sec.Uncertainty-Aware Fusion}
Given the RoI features $F_{\text{cam}}$ and $F_{\text{lid}}$ from two modalities, the Uncertainty-Aware Fusion (UAF) module aims to endow our fusion module with the robustness against sensor noise illustrated in \cref{fig.UAF}. To this end, we inject the awareness of the uncertainty of each modality into our fusion module, \ie,
\begin{equation}
  \bar{Q} = f_{UA}(F_{\text{cam}}, U_{\text{cam}}, F_{\text{lid}},  U_{\text{lid}}),
  \label{eq.F}
\end{equation}
where $\bar{Q}=\{\bar{q}_i\}_{i=1}^{N_q} \subset \mathbb{R}^{C}$ and $f_{UA}$ are the refined query feature and uncertainty-aware fusion function, respectively. Besides, $U_{\text{cam}}$ and $U_{\text{lid}}$ are the uncertainty of two modalities. 

Inspired by the unquestionable importance of accurate localization in autonomous driving, we formulate the uncertainty as a function of the Euclidean distance between the predicted and the ground-truth bounding boxes $B$. For convinience, let $s \in \{\text{cam}, \text{lid}\}$ represents one modality. We have  
\begin{equation}
  U_{s} = 1-exp\left(-D^{xy}\left(f_{\text{reg}}(F_{\text{s}}), B\right)\right),
  \label{eq.u}
\end{equation}
where $f_{\text{reg}}$ is the regression function for bounding boxes, and $D^{xy}$ is the Euclidean distance function in BEV space. However, the ground-truth bounding boxes are unavailable for models. Thus, we inject a distance predictor on RoI features of each modality, then rewrite \cref{eq.u} as 
\begin{equation}
  \hat{U}_{s} = 1-exp\left(-f_{\text{dist}}(F_{\text{s}})\right),
\end{equation}
where $f_{\text{dist}}$ is the distance predictor consisting of MLPs. 

As for the uncertainty-aware fusion function $f_{UA}$, we simply formulate it as the concatenation of features weighted by uncertainty and rewrite \cref{eq.F} as  
\begin{equation}
\bar{q}_i = FFN\left(Cat\left( F_{\text{cam}} \cdot (1- \hat{U}_{\text{cam}}) , F_{\text{lid}}   \cdot (1-\hat{U}_{\text{lid}})\right)\right),
  \label{eq.F2}
\end{equation}
where $Cat$ and $FFN$ denote concatenation function and feedforward networks, respectively. In this way, our UAF module quantifies the uncertainty $U$ of each modality and guides our model to focus on trustworthy modality, thus enjoying robustness against sensor noises. 

\begin{table}[tb]
  \caption{Quantitative comparisons of SparseLIF with all state-of-the-art 3D detectors on the nuScenes test benchmark. The notion of modality: Camera (C), LiDAR (L), and Temporal (T). $\dagger$: using external training data; $\ddagger$: using TTA and complex model ensemble (\eg models with different voxel sizes, BEV sizes, backbones/FPNs/heads); $\S$: we only use very simple self-model ensemble without TTA for \emph{SparseLIF-T}.
} 
  \label{Tab.test}
  \centering
  \resizebox{1.0\linewidth}{!}{ 
\begin{tabular}{l|c|ccccc|cc}
\toprule
Method & Modality & mATE$\downarrow$ & mASE$\downarrow$ & mAOE$\downarrow$ & mAVE$\downarrow$ & mAAE$\downarrow$ & mAP$\uparrow$ & \textcolor{red}{NDS$\uparrow$} \\ \midrule
TransFusion~\cite{bai2022transfusion} & LC & 25.9 & 24.3 & 35.9 & 28.8 & 12.7 & 68.9 & 71.7 \\
FUTR3D~\cite{chen2023futr3d} & LC & 28.4 & 24.1 & 31.0 & 30.0 & 12.0 & 69.4 & 72.1 \\
AutoAlignV2~\cite{chen2022autoalignv2} & LC & 24.5 & 23.3 & 31.1 & 25.8 & 13.3 & 68.4 & 72.4 \\
BEVFusion~\cite{liu2023bevfusion} & LC & 26.1 & 23.9 & 32.9 & 26.0 & 13.4 & 70.2 & 72.9 \\
BEVFusion~\cite{liang2022bevfusion} & LC & 25.0 & 24.0 & 35.9 & 25.4 & 13.2 & 71.3 & 73.3 \\
DeepInteraction~\cite{yang2022deepinteraction} & LC & 25.7 & 24.0 & 32.5 & 24.5 & 12.8 & 70.8 & 73.4 \\
BEVFusion4D-S~\cite{cai2023bevfusion4d} & LC & \multicolumn{1}{c}{-} & \multicolumn{1}{c}{-} & \multicolumn{1}{c}{-} & \multicolumn{1}{c}{-} & - & 71.9 & 73.7 \\
SparseFusion~\cite{zhou2023sparsefusion} & LC & \multicolumn{1}{c}{-} & \multicolumn{1}{c}{-} & \multicolumn{1}{c}{-} & \multicolumn{1}{c}{-} & - & 72.0 & 73.8 \\
MSMDFusion~\cite{jiao2023msmdfusion} & LC & 25.5 & 23.8 & 31.0 & 24.4 & 13.2 & 71.5 & 74.0 \\
CMT~\cite{yan2023cross} & LC & 27.9 & 23.5 & 30.8 & 25.9 & 11.2 & 72.0 & 74.1 \\
EA-LSS~\cite{hu2023ea} & LC & 24.7 & 23.7 & 30.4 & 25.0 & 13.3 & 72.2 & 74.4 \\
UniTR~\cite{wang2023unitr} & LC & 24.1 & 22.9 & \textbf{25.6} & 24.0 & 13.1 & 70.9 & 74.5 \\
FocalFormer3D-F~\cite{chen2023focalformer3d} & LC & 25.1 & 24.2 & 32.8 & 22.6 & 12.6 & 72.4 & 74.5 \\
BEVFusion4D~\cite{cai2023bevfusion4d} & LCT & - & \multicolumn{1}{c}{-} & \multicolumn{1}{c}{-} & \multicolumn{1}{c}{-} & - & 73.3 & 74.7 \\
DAL~\cite{huang2023detecting} & LC & 25.3 & 23.8 & 33.4 & 17.4 & 12.0 & 72.0 & 74.8 \\
FusionFormer$^\dagger$~\cite{hu2023fusionformer} & LCT & 26.7 & 23.6 & 28.6 & 22.5 & \textbf{10.5} & 72.6 & 75.1 \\
\textbf{\emph{SparseLIF-T}} & LCT & \textbf{24.1}  & \textbf{22.9} & 27.8 & \textbf{15.4} & 11.8 & \textbf{74.4}  & \textbf{\textcolor{red}{77.0}}  \\ \midrule
PAI3D$^{\ddagger}$~\cite{liu2022pai3d} & LC & 24.5 & 23.3 & 30.8 & 23.3 & 13.1 & 71.4 & 74.2 \\
Lift-Attend-Splat$^{\ddagger}$~\cite{gunn2023lift} & LC & 24.3 & 23.8 & 34.5 & 32.8 & 13.3 & 75.5 & 74.9 \\
BEVFusion$^{\ddagger}$~\cite{liu2023bevfusion} & LC & 24.2 & 22.7 & 32.0 & 22.2 & 13.0 & 75.0 & 76.1 \\
DeepInteraction$^{\ddagger}$~\cite{yang2022deepinteraction} & LC & 23.5 & 23.3 & 32.8 & 22.6 & 13.0 & 75.6 & 76.3 \\
CMT$^{\ddagger}$~\cite{yan2023cross} & LC & 23.3 & \textbf{22.0} & \textbf{27.1} & 21.2 & 12.7 & 75.3 & 77.0 \\
BEVFusion4D$^{\ddagger}$~\cite{cai2023bevfusion4d} & LCT & \textbf{22.9} & 22.9 & 30.2 & 22.5 & 13.5 & \textbf{76.8} & 77.2 \\
EA-LSS$^{\ddagger}$~\cite{hu2023ea} & LC & 23.4 & 22.8 & 27.8 & 20.4 & 12.4 & 76.6 & 77.6 \\
\textbf{\emph{SparseLIF-T}}$^{\S}$ & LCT & 24.3 & 23.1 & 28.4 & \textbf{15.2} & \textbf{11.7} & 75.9 & \textbf{\textcolor{red}{77.7}} \\ \bottomrule
\end{tabular}
    }
\end{table}

\begin{table}[tb]
  \caption{Quantitative comparisons of SparseLIF with all state-of-the-art 3D detectors on the nuScenes validation set. The notion of modality: Camera (C), LiDAR (L), and Temporal (T). $\dagger$: with extra CBGS training strategy. Note that all methods use corresponding best single model without TTA or model ensemble for comparisons.
}
  \label{Tab.val}
  \centering
  \vspace{-0.1cm}\vspace{-0.1cm}
\begin{tabular}{l|c|cc|cc}
\toprule
    Method & Modality & Image Backbone & LiDAR Backbone  & mAP$\uparrow$ & \textcolor{red}{NDS$\uparrow$} \\ \midrule
FUTR3D~\cite{chen2023futr3d} & LC & ResNet-101 & VoxelNet & 64.2 & 68.0 \\
AutoAlignV2~\cite{chen2022autoalignv2} & LC & CSPNet & VoxelNet & 67.1 & 71.2 \\
TransFusion~\cite{bai2022transfusion} & LC & ResNet-50 & VoxelNet & 67.5 & 71.3 \\
BEVFusion~\cite{liu2023bevfusion} & LC & SwinT & VoxelNet & 68.5 & 71.4 \\
BEVFusion~\cite{liang2022bevfusion} & LC & SwinT & VoxelNet & 69.6 & 72.1 \\
DeepInteraction~\cite{yang2022deepinteraction} & LC & ResNet-50 & VoxelNet & 69.9 & 72.6 \\
CMT~\cite{yan2023cross} & LC & V2-99 & VoxelNet & 70.3 & 72.9 \\
BEVFusion4D-S~\cite{cai2023bevfusion4d} & LC & SwinT & VoxelNet & 70.9 & 72.9 \\
SparseFusion~\cite{zhou2023sparsefusion} & LC & SwinT & VoxelNet & 71.0 & 73.1 \\
EA-LSS~\cite{hu2023ea} & LC & SwinT & VoxelNet & {71.2} & 73.1 \\
FusionFormer-S$^\dagger$~\cite{hu2023fusionformer} & LC & V2-99 & VoxelNet & 70.0 & {73.2} \\
\textbf{\emph{SparseLIF-S}} & LC & V2-99 & VoxelNet & \textbf{71.2} & \textbf{\textcolor{red}{74.6}} \\  \midrule
BEVFusion4D~\cite{cai2023bevfusion4d} & LCT & SwinT & VoxelNet & {72.0} & 73.5 \\
FusionFormer$^\dagger$~\cite{hu2023fusionformer} & LCT & V2-99 & VoxelNet & 71.4 & {74.1} \\
\textbf{\emph{SparseLIF-T}} & LCT & V2-99 & VoxelNet & \textbf{74.7} & \textbf{\textcolor{red}{77.5}} \\ \bottomrule
\end{tabular}
\end{table}

\begin{table}[tb]
  \caption{Ablation studies of SparseLIF on the nuScenes validation set. }
  \label{Tab.ablation}
  \centering
  \vspace{-0.1cm}\vspace{-0.1cm}
  \resizebox{1.0\linewidth}{!}{ 
\begin{tabular}{c|c|cc|ccc|cc}
\toprule
\emph{SparseLIF-S} & Modality & Image Backbone & LiDAR Backbone & PAQG & RIAS & UAF & mAP$\uparrow$ & NDS$\uparrow$ \\ \midrule
\#1 & LC & V2-99 & VoxelNet & &  &  & 66.2 & 69.0 \\
\#2 & LC & V2-99 & VoxelNet & & \checkmark &  & 69.8 & 73.3 \\
\#3 & LC & V2-99 & VoxelNet & \checkmark & \checkmark &  & {71.0} & {74.3} \\
\#4 & LC & V2-99 & VoxelNet & & \checkmark & \checkmark  & 70.5 & 74.1 \\
\#5 & LC & V2-99 & VoxelNet & \checkmark & \checkmark & \checkmark & \textbf{71.2} & \textbf{74.6} \\ 
\#6 & LC & V2-99 & VoxelNet & \checkmark &  & \checkmark & 68.0 & 70.8 \\
\bottomrule
\end{tabular}
    }
\end{table}

\begin{table}[tb]
\vspace{-0.1cm}
  \caption{Performance analysis of our PAQG module on detection distances and small object classes on the nuScenes validation set, based on \emph{SparseLIF-S}. The $AP$ scores of traffic cone (T.C.) and barrier at $30m$- are missing since corresponding annotations are unavailable.}
  \label{Tab.PAQG}
  \centering
  \vspace{-0.1cm}\vspace{-0.1cm}
\begin{tabularx}{10cm}{p{1.2cm}|p{1.2cm}<{\centering}|p{1.8cm}<{\centering}p{1.8cm}<{\centering}p{1.8cm}<{\centering}p{1.8cm}<{\centering}}
\toprule
 & PAQG & $0$-$10m$ & $10$-$20m$ & $20$-$30m$ & $30m$- \\ \toprule
mAP &  & 75.1 & 74.3 & 65.5 & 56.9 \\
mAP & \checkmark & \textbf{76.2} & \textbf{74.8} & \textbf{66.8} & \textbf{58.5} \\ \midrule
T.C. &  & 82.5 & 80.6 & 69.1 & - \\
T.C. & \checkmark & \textbf{83.8} & \textbf{82.7} & \textbf{70.4} & - \\ \midrule
Barrier &  & 72.1 & 77.5 & 54.5 & - \\
Barrier &\checkmark & \textbf{76.4} & \textbf{80.0} & \textbf{63.1} & - \\ \bottomrule
\end{tabularx}
    \vspace{-0.1cm}\vspace{-0.1cm}\vspace{-0.1cm}
\end{table}

\section{Experiments} \label{sec.Experiments}
This section provides the experimental settings and results. We conduct detailed ablation studies to verify our design choices in SparseLIF. Meanwhile, we also demonstrate that our multi-modality detector achieves excellent robustness against sensor noises. Above all, we compare SparseLIF with other state-of-the-art 3D object detectors on the popular nuScenes benchmark. The results show that our SparseLIF achieves superior performance, ranking \textbf{\emph{1st}} on both the validation set and test benchmark.

\subsection{Experimental Setups}

\subsubsection{Implementation Details} 
We implement SparseLIF using the open-source MMDetection3D~\cite{mmdet3d2020} based on PyTorch. The detection range is $[-54m,54m]$ and $[-5m,3m]$ for the XY- and the Z-axis. We adopt V2-99~\cite{lee2020centermask} pretrained by FCOS3D~\cite{wang2021fcos3d} as the image backbone with input image size $1600\times640$. We adopt VoxelNet~\cite{zhou2018voxelnet} as LiDAR backbone with voxel size $(0.075m,0.075m,0.2m)$. The total query number $N_q$ is $900$, including $N_k=200$ queries generated by the PAQG module. The perspective detector is implemented by the coupled FCOS~\cite{tian2019fcos} and FCOS3D~\cite{wang2021fcos3d} sub-networks. The lightweight distance predictor $f_{dist}$ is implemented by a two-layer FFN. The decoder repeats $L=6$ times. In the following experiments, we report the state-of-the-art performance of two SparseLIF detectors: the single-frame detector \emph{\textbf{SparseLIF-S}} ($V=6$, $M=4$, $R=4$, and $T=1$), the temporal multi-frame detector \emph{\textbf{SparseLIF-T}} ($V=6$, $M=4$, $R=4$, and $T=13$).


Each model is trained end-to-end using the AdamW optimizer on eight NVIDIA A100 GPUs with a total batch size of $8$. For fair comparisons, we apply the query-denoising strategy~\cite{li2022dn}, commonly used in sparse detection heads, to address the unstable matching problem. Each model is trained for $24$ epochs with a learning rate of $2e-4$.

\subsubsection{Datasets and Evaluation Metrics} We conduct experiments on the popular nuScenes dataset~\cite{caesar2020nuscenes} to evaluate the performance of our proposed method for 3D object detection in autonomous driving. The nuScenes dataset has $1.4$ million 3D detection annotation boxes from $40,157$ samples distributed in $1000$ scenes collected in Boston and Singapore. Each sample is collected with six cameras and a $32$-beam LiDAR sensor. We adopt the nuScenes detection evaluation metrics $NDS$ and $mAP$ over ten classes for our experiments.

\subsection{Comparisons with State-of-the-Art 3D Object Detectors}
As shown in the top part of \cref{Tab.test}, without using any test-time augmentation (TTA) or model ensemble, our \emph{SparseLIF-T} achieves state-of-the-art single-model performance, reaching $77.0\%$ $NDS$ on the nuScenes test benchmark, significantly outperforming all other 3D detectors by a notable margin. In particular, we outperform the most competitive method FusionFormer~\cite{hu2023fusionformer} by $1.9\%$ $NDS$ without using any external training data. Regarding the test benchmark leaderboard in the bottom part of \cref{Tab.test}, many competitive methods~\cite{liu2022pai3d,gunn2023lift,liu2023bevfusion,yang2022deepinteraction,yan2023cross,cai2023bevfusion4d,hu2023ea} adopt very complex model ensemble (\eg assembling models with different voxel sizes, BEV sizes, backbones/FPNs/heads) and TTA, to strive for top ranking on the test leaderboard. Contrarily, we only use very simple self-model ensemble without TTA for \emph{SparseLIF-T} (\ie, $0.7\%$ $NDS$ improvement), achieving the best performance of $77.7\%$ $NDS$ and ranking \textbf{\emph{1st}} on the test leaderboard by the time of paper submission.

SparseLIF is one of the first LiDAR-camera-based 3D detectors~\cite{cai2023bevfusion4d, hu2023fusionformer} with temporal awareness, while most methods are ignorant or incapable of integrating temporal information, resulting in sub-optimal performance. For fair comparisons, we also compare our single-frame detector \emph{SparseLIF-S} with other temporal-ignorant methods on the nuScenes validation set. As shown in the top part of \cref{Tab.val}, \emph{SparseLIF-S} also outperforms the best competitor by a notable margin ($1.4\%$ $NDS$). Furthermore, as presented in the bottom part of \cref{Tab.val}, our multi-frame detector \emph{SparseLIF-T} achieves the $NDS$ of $77.5\%$, significantly outperforming all other methods by at least $3.4\%$ on the validation set. 

We also conduct latency analysis on the nuScenes dataset. We implement SparseLIF using Pytorch without any acceleration operations. The overall latency of \emph{SparseLIF-S} is $340ms$ on a single NVIDIA A100 GPU. In detail, the detection head (including the PAQG module, the RIAS module, and the UAF module, \etc) only takes about $40ms$, while the camera and LiDAR backbones take the rest of the time, which demonstrates the efficiency of our detector. We can further speed up our detector by configuring the backbones.

\begin{table}
  \caption{Robustness studies of SparseLIF on the nuScenes validation set, under challenging scenarios: LiDAR malfunction, camera malfunction and unsynchronization. Abbreviations: construction vehicle (C.V.), pedestrian (Ped.), and traffic cone (T.C.). 
}
  \label{Tab.robust}
  \centering
  \resizebox{1.0\linewidth}{!}{ 
\begin{tabular}{ll|c|cccccccccc|cc}
\toprule
\multicolumn{2}{l|}{Setting}                     & UAF & Car  & Truck & Bus  & Trailer & C.V. & Ped. & Motor & Bike & T.C. & Barrier & mAP  & NDS  \\ \midrule
\multicolumn{1}{l|}{\multirow{4}{*}{FOV}} &\multicolumn{1}{c|}{\multirow{2}{*}{$120^{\circ}$}} &  & 69.4 & 49.9 & 62.6 & 23.2 & 20.0 & 56.6 & 55.7 & 50.4 & 53.2 & 55.4 & 49.7 & 62.1 \\
 \multicolumn{1}{c|}{}& & \checkmark & \textbf{74.5} & \textbf{55.1} & \textbf{66.5} & \textbf{33.9} & \textbf{21.7} & \textbf{60.1} & \textbf{61.3} & \textbf{57.3} & \textbf{60.6} & \textbf{62.7} & \textbf{55.4} & \textbf{65.2} \\  \cline{2-15} 
\multicolumn{1}{c|}{}& \multicolumn{1}{c|}{\multirow{2}{*}{$180^{\circ}$}} &  & 73.6 & 55.9 & 64.2 & 27.3 & 22.8 & 65.0 & 61.1 & 54.8 & 59.0 & 59.0 & 54.3 & 65.3 \\
\multicolumn{1}{c|}{} && \checkmark & \textbf{77.5} & \textbf{59.6} & \textbf{68.5} & \textbf{35.8} & \textbf{23.1} & \textbf{67.7} & \textbf{64.6} & \textbf{60.3} & \textbf{65.4} & \textbf{65.1} & \textbf{58.8} & \textbf{67.6} \\ \midrule
\multicolumn{2}{l|}{\multirow{2}{*}{\shortstack{Object\\Failure}}} &  & 84.9 & 64.2 & 74.4 & 37.8 & 27.9 & 82.1 & 75.5 & 70.3 & 74.1 & \textbf{70.1} & 66.1 & 72.2 \\
& & \checkmark & \textbf{86.2} & \textbf{67.4} & \textbf{75.2} & \textbf{41.6} & \textbf{29.2} & \textbf{82.4} & \textbf{76.4} & \textbf{71.4} & \textbf{75.4} & 69.8 & \textbf{67.5} & \textbf{73.0} \\ \midrule \midrule
\multicolumn{2}{l|}{Front} &  & 82.2 & 59.9 & 65.3 & 35.3 & 29.0 & \textbf{86.4} & 64.8 & 69.1 & 78.2 & 66.2 & 63.7 & 71.4 \\
\multicolumn{2}{l|}{Occlusion} & \checkmark & \textbf{84.1} & \textbf{62.8} & \textbf{66.7} & \textbf{36.5} & \textbf{30.3} & 84.8 & \textbf{66.5} & \textbf{70.2} & \textbf{78.5} & \textbf{66.3} & \textbf{64.7} & \textbf{72.0} \\ \midrule \midrule
\multicolumn{2}{l|}{\multirow{2}{*}{Stuck}} &  & 90.2 & 69.9 & 81.5 & 43.8 & 33.0 & 90.7 & 82.9 & \textbf{77.1} & 82.6 & 74.9 & 72.7 & 75.9 \\
& & \checkmark & \textbf{90.7} & \textbf{72.3} & \textbf{82.0} & \textbf{46.3} & \textbf{33.1} & \textbf{90.8} & \textbf{83.9} & 76.9 & \textbf{83.0} & \textbf{75.0} & \textbf{73.4} & \textbf{76.5} \\ \bottomrule
\end{tabular}
}
\end{table}

\begin{figure}[tb]
  \centering
  \includegraphics[width=\linewidth]{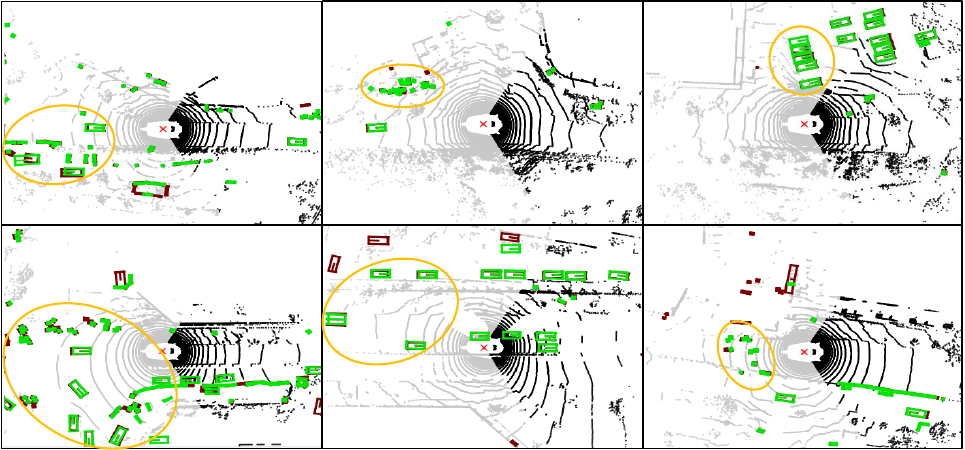}
  \caption{Robustness visualizations under the scenario of limited LiDAR FOV angle of $120^{\circ}$. We color each box with \textcolor{mygreen}{green} and \textcolor{myred}{red} for prediction and ground truth.  
    }
  \label{fig.bbox}
\end{figure}

\subsection{Ablation Studies}
In \cref{Tab.ablation}, we conduct ablation studies on the nuScenes validation set to evaluate the key components in our multi-modality framework, based on the state-of-the-art single-frame detector \emph{SparseLIF-S}, which yields highly convincing proofs. The RIAS module plays a vital role in our multi-modality detector. Adopting the PAQG module to generate high-quality query proposals, the $mAP$ and $NDS$ are improved by $1.2\%$ and $1.0\%$ respectively. Adopting the UAF module to conduct multi-modality fusion, the $mAP$ and $NDS$ are improved by $0.7\%$ and $0.8\%$ respectively. When \emph{SparseLIF-S} assembles all modules, the best performance is reached: $71.2\%$ $mAP$ and $74.6\%$ $NDS$. 

We further present an in-depth analysis of our proposed PAQG module to reveal its effectiveness on detection distances and small object classes, based on the state-of-the-art single-frame detector \emph{SparseLIF-S}. As shown in \cref{Tab.PAQG}, the PAQG module substantially facilitates distant object detection, \eg $1.6\%$ $mAP$ improvement for objects beyond $30m$. Regarding small objects, the PAQG module also significantly improves the $AP$ scores of traffic cone and barrier across all distances, \eg $8.6\%$ $AP$ improvement for barrier in $20$-$30m$.
We attribute the performance gains to the enhanced awareness of rich context and perspective priors boosted by the proposed PAQG module. 

\subsection{Robustness Studies}
To validate the robustness of our multi-modality framework, we evaluate SparseLIF under LiDAR/camera malfunction and unsynchronization scenarios (see~\cite{yu2023benchmarking} for more details): 
\begin{itemize}
    \item \textbf{Limited FOV}. We simulate the limited FOV angles of $120^{\circ}$ and $180^{\circ}$ by filtering out LiDAR points.
    \item \textbf{Object Failure}. Following BEVFusion~\cite{liang2022bevfusion}, we simulate this scenario by selecting $50\%$ frames to drop points of objects, where $50\%$ objects are dropped for each selected frame. 
    \item \textbf{Front Occlusion}. Following BEVFusion~\cite{liang2022bevfusion}, we simulate such an occlusion scenario by filling the entire front-camera image with zero value.
    \item \textbf{Stuck}. The timestamps of two sensors might not always be synchronized, yielding stuck data, \eg the detector wrongly receives data with timestamp $t-1$ at time $t$. Following BEVFusion~\cite{liang2022bevfusion}, we simulate such an unsynchronized scenario on $50\%$ frames.
\end{itemize}
We directly evaluate our \emph{SparseLIF-T} model under these scenarios without any adaption or fine-tuning. As shown in \cref{Tab.robust}, the UAF module boosts the robustness performance by $3.1\%$ $NDS$ at the most challenging LiDAR malfunction scenario (top), \ie, limited FOV angle of $120^{\circ}$.
Simultaneously, our SparseLIF also gains robustness improvement by $0.6\%$ $NDS$ under camera malfunction (middle) and unsynchronization (bottom) scenarios. The experimental results convincingly demonstrate the capability of our detector against sensor noises. 

We further visualize the predictions of our SparseLIF under the most challenging scenario of limited LiDAR FOV angle of $120^{\circ}$. As presented in \cref{fig.bbox}, our SparseLIF precisely detects objects in golden circles with LiDAR input malfunctioned, showing the remarkable robustness of our multi-modality detector attributed to the proposed UAF module. 


\section{Conclusion} \label{sec.Conclusion}
This paper proposes a high-performance fully sparse detector termed SparseLIF for LiDAR-camera-based 3D object detection. Our SparseLIF achieves state-of-the-art performance by enhancing the awareness of rich representations in two modalities. In particular, SparseLIF consists of (1) the PAQG module, which generates high-quality 3D queries with perspective priors to facilitate the perception of small and distant objects; (2) the RIAS module, which further refines prior queries by RoI feature sampling to embrace the fully sparse paradigm with the capability of low latency and integration of more temporal frames; (3) the UAF module, which quantifies the uncertainty of each modality for multi-modality fusion to enhance robustness against sensor noises. The experimental results demonstrate the superiority of our proposed method over all state-of-the-art 3D object detectors on the nuScenes benchmark. In the future, we will explore applications of SparseLIF on other tasks, such as occupancy prediction.


%
%
\bibliographystyle{splncs04}
\bibliography{main}
\end{document}